\documentclass{article}
\usepackage{amsmath,epsfig}
\usepackage[preprint]{spconfa4}
\usepackage{xcolor}
\usepackage{cite}

\usepackage{amsfonts}
\usepackage{color}

\let\OLDthebibliography\thebibliography
\renewcommand\thebibliography[1]{
  \OLDthebibliography{#1}
  \setlength{\parskip}{0pt}
  \setlength{\itemsep}{0pt plus 0.3ex}
}

\begin{document}\sloppy

\def\x{{\mathbf x}}
\def\L{{\cal L}}

\title{Deep Point Cloud Normal Estimation via Triplet Learning}
%
\name{Weijia Wang, Xuequan Lu, Dasith de Silva Edirimuni,  Xiao Liu, Antonio Robles-Kelly}
\address{School of Information Technology, Deakin University, \\
Waurn Ponds Campus, Geelong, VIC 3216, Australia\\
\{wangweijia, xuequan.lu, dtdesilva, xiao.liu, antonio.robles-kelly\}@deakin.edu.au}

\maketitle

\begin{abstract}
Current normal estimation methods for 3D point clouds often show limited accuracy in predicting normals at sharp features (e.g., edges and corners) and less robustness to noise. In this paper, we propose a novel normal estimation method for point clouds which consists of two phases: (a) feature encoding to learn representations of local patches, and (b) normal estimation that takes the learned representation as input and regresses the normal vector. We are motivated that local patches on isotropic and anisotropic surfaces respectively have similar and distinct normals, and these separable features or representations can be learned to facilitate normal estimation. To realise this, we design a triplet learning network for feature encoding and a normal estimation network to regress normals. Despite having a smaller network size compared with most other methods, experiments show that our method preserves sharp features and achieves better normal estimation results especially on computer-aided design (CAD) shapes.
\end{abstract}

\section{Introduction}
\label{sec:intro}


Point cloud data is a representation of 3D geometry which has been applied in a wide range of fields, including autonomous driving, robotics and augmented reality~\cite{Lu_Pointfilter_2021, Yew_2018}. Raw point clouds consist of unordered points that lack normal information, and are usually corrupted by noise due to limitations on the scanning devices' precision~\cite{Lu_DFP_2020, Guerrero_pcpnet_2018, Benshabat_Nesti_2019, Qi_2017, Hoppe_1992}. This makes it challenging to directly use raw point clouds for visual computing tasks such as surface reconstruction, shape smoothing and segmentation~\cite{Lu_Pointfilter_2021,Benshabat_Nesti_2019}. Estimating reliable normal information on the input point clouds has proven to be significant in achieving satisfactory results in such tasks. 

Conventional point cloud normal estimation methods based on principal component analysis (PCA)~\cite{Hoppe_1992} have limited accuracy in estimating normals at sharp features and are sensitive to noise. In recent years, with the development of deep learning for point clouds, learning-based normal estimation methods have been proposed to improve performance in this area, but there are still limitations among them. For example, the methods proposed in~\cite{Boulch_Hough_2016, Guerrero_pcpnet_2018, Wang_NINormal_2020} do not perform well in the presence of noise and in predicting normals at sharp features. The work presented in~\cite{Lu_DFP_2020} shows more robust performance on noisy inputs yet consists of a two-step testing phase that requires a long time to produce results. Nesti-Net~\cite{Benshabat_Nesti_2019}, while being a promising candidate, has a large network size and lengthy inference time.

In this paper, we introduce a novel normal estimation method for point clouds with triplet learning to address those limitations. We are motivated to exploit a triplet learning network that brings representations (or features) of similar point cloud patches close to each other, and pushes representations of dissimilar patches away from each other, which facilitates effective normal estimation. We employ PointNet~\cite{Qi_2017} as the feature-extraction backbone and treat the local neighbourhood of each point (i.e., central point) as a local patch. To define a triplet, we first select a local patch as the anchor, and regard another local patch with a small angle difference between their central point normals as the positive sample, and the third patch with a large angle difference as the negative sample. We derive this based on the fact that two local patches on the same isotropic surface should have similar central normals and representations. This phase learns patch features which will be used as input for the normal estimation phase. We design multilayer perceptrons (MLPs) as the normal estimation network and use it to regress the normal vector from the encoded features of a local patch, trained by our designed cosine similarity loss function.

Our training and testing point cloud datasets consist of both CAD and non-CAD synthetic shapes, where CAD shapes contain sharp features while non-CAD shapes generally have smoother surfaces. Experiments show that our method achieves better results than other normal estimation techniques especially on noisy CAD shapes, while comparable performance is gained on smoother non-CAD shapes. Our method has several advantages: (a) it performs very well on noisy CAD shapes (both synthetic and scanned) in terms of preserving sharp features, (b) it is robust to irregular sampling and fewer points, (c) it has a small network size of 10.42 MB and it completes normal estimation within a short amount of time (i.e., 55.6 seconds per 100,000 points), and (d) the single trained network can estimate normals for both CAD and non-CAD shapes. 


\section{Related Work}
\label{sec:related-work}

\subsection{Normal Estimation for Point Clouds}

As a commonly adopted traditional normal estimation approach, PCA~\cite{Hoppe_1992} derives each point's normal by calculating the eigenvalues and eigenvectors of the covariance matrix of its neighbourhood. Nevertheless, PCA-based method and its variants tend to smooth out sharp edges and corners that exist in the input shapes. In recent years, learning-based normal estimation methods have started to emerge, but only a few of these methods showcase the ability to preserve sharp features, handle noise robustly and generalise among different shapes. As the pioneer of such approaches, Boulch and Marlet~\cite{Boulch_Hough_2016} present HoughCNN, in which 3D point clouds are mapped to a 2D Hough space and, thereafter, a convolutional neural network (CNN) performs normal estimation on this representation. However, this transformation to the 2D space may discard important geometrical details. After the introduction of PointNet~\cite{Qi_2017}, 3D points can be directly fed into networks and the feature representations can be directly extracted without being affected by the input sequence. Adopting this as the backbone, PCPNet~\cite{Guerrero_pcpnet_2018} encodes local neighbourhoods on point clouds as patches, and regresses normals from them. Nesti-Net~\cite{Benshabat_Nesti_2019} adopts a mixture-of-experts network that checks points' neighbourhoods in varying scales and selects the optimal one. Similarly, NINormal~\cite{Wang_NINormal_2020} adopts an attention module that softly selects neighbouring points to regress normals from them. To effectively preserve feature normals, Lu et al.~\cite{Lu_DFP_2020} classify points into feature and non-feature classes, and perform denoising on noisy point clouds in conjunction with normal estimation in their Deep Feature Preserving (DFP) method. Recently, Lenssen et al.~\cite{Lenssen_DI_2020} proposed Deep Iterative (DI), a graph neural network-based approach that performs a weighted least squares plane-fitting of point neighbourhoods, where the weights are iteratively refined during the process.

\subsection{Triplet Loss}
The triplet loss function, which focuses on optimizing the relative distances of an anchor to a positive and negative sample, has been applied to a broad range of contexts. Traditionally, triplet loss was mainly applied to 2D image processing tasks, such as image learning~\cite{Huang_crossdomain_2015}, face identification~\cite{Schroff_2015} and person matching~\cite{Zhang-inc-tri-2019}. Extending from 2D, triplet loss has also been employed for 3D geometric data processing in recent years. For example,~\cite{Yew_2018} and~\cite{Deng_2018} use triplet loss for point cloud matching, and~\cite{Cheraghian_TZSL_2020} employs unsupervised triplet loss for point cloud classification. While triplet loss has proven to be effective for feature-learning tasks on 3D data, it has never been extended to normal estimation on 3D point clouds.

\section{Method}
\label{sec:method}

\begin{figure}[t]
  \centering

  \includegraphics[width=\columnwidth]{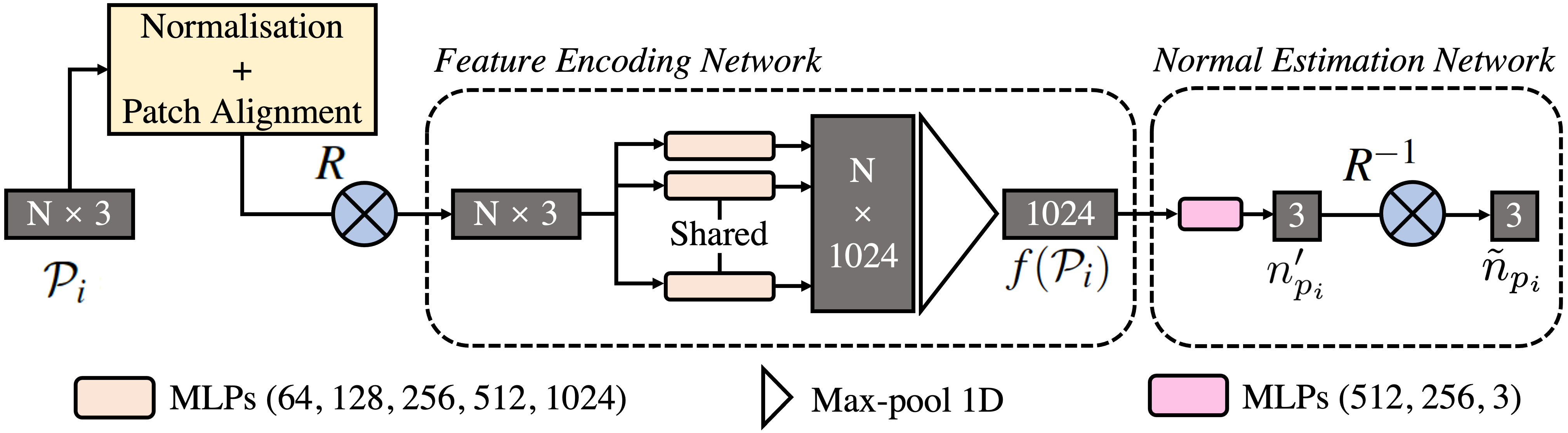}
  \caption{The overall architecture of our method, where the patch size N is empirically set to 500.
  }
  \label{fig:overall_pipeline}
\end{figure}

Our normal estimation approach consists of two phases: (a) feature encoding that learns geometric features from input point clouds, and (b) normal estimation that regresses normals from the encoded features. In phase (a), we first define the local patch as a point (i.e., central point) with its neighbouring points, and conduct pre-processing to mitigate pose and point number inconsistencies. We then construct triplets of patches and feed them to the feature encoding network (i.e., a triplet network). In phase (b), the latent representations learned by phase (a) are consumed by the normal estimation network which outputs a predicted normal for the given patch (i.e., normal for the central point of that patch). Note that a rotation matrix is used in phase (a) for patch alignment, and its inverse matrix is further used in phase (b) for recovering back to the original space. Fig.~\ref{fig:overall_pipeline} shows the overall architecture of our method, which is trained in a supervised manner. 


\subsection{Feature Learning} 
\label{sec:feature-encoder}

\label{sec:pre-process}
\textbf{Patch Pre-processing.} Before training, we pre-process each point cloud into patches as inputs for our model. For any point cloud $\mathbf{P} = \{p_1,\ldots, p_n\ |\ p_i \in \mathbb{R}^3,\ i\ = 1, ..., n \}$, we define a local patch $\mathcal{P}_i$ centered at a point $p_i$ as:
\begin{equation} 
\label{eq:patch}
    \mathcal{P}_i = \{ p_j\ |\ \left \| p_j - p_i \right \| < r \},
\end{equation}
where $p_j$ is any point within the patch $\mathcal{P}_i$ and $r$ is the radius of the ball centered at point $p_i$. 

There are two issues associated with raw patches, making them unsuitable for effective learning: (a) they contain unnecessary degrees of freedom and rigid transformations, and (b) each patch's number of points might vary, which prohibits them from being grouped into input batches during the training phase~\cite{Guerrero_pcpnet_2018, Lu_Pointfilter_2021}. To address (a), we first translate each patch to its origin and normalise its size, such that $\mathcal{P}_i = ( \mathcal{P}_i - p_i) / r$. Inspired by~\cite{Lu_Pointfilter_2021}, we then align each patch's last principal axis with the $z$-axis and the second principal axis with the $x$-axis in Cartesian space using a rotation matrix $R$, computed using the principal components of the patch's covariance matrix. As a result, the unnecessary degrees of freedom are removed, leaving the patch invariant under rigid transformations; the inverse rotation matrix $R^{-1}$ is required later, during testing. To solve (b), we select $k$ points from each patch to ensure the input patches are of a consistent size. For raw patches with more points than $k$, we randomly screen $k$ points; otherwise, we pad the input using existing points within the patch to make up $k$ points. We empirically set $k$ to be 500 and the ball radius $r$ to be 5\% of the point cloud's bounding box diagonal. 

\textbf{Triplet Generation.} As defined in~\cite{Schroff_2015}, each data sample is treated as an anchor and needs to be paired with a positive and a negative sample in order to construct a triplet. We treat each input patch as an anchor patch $\mathcal{P}_i$, which is paired with a positive and negative patch. We denote the ground-truth normal of the anchor patch's central point, $p_i$, by $n_{p_{i}}$. Thereafter, we define a positive patch $\mathcal{S}_i$ and negative patch $\mathcal{T}_i$ according to the anchor, as:
\begin{align} 
\mathcal{S}_i &= \{ s_j\ |\ \left \| s_j - s_i \right \| < r,\ \theta(n_{p_i}, n_{s_i}) \leq \theta_{th} \}, \\
\mathcal{T}_i &= \{ t_{j}\ |\ \left \| t_{j} - t_{i} \right \| < r,\ \theta(n_{p_{i}}, n_{t_{i}}) > \theta_{th} \},
\end{align}
where $s_{i}$ and $t_{i}$ are respectively the central points of $\mathcal{S}_i$ and $\mathcal{T}_i$, and $n_{s_{i}}$ and $n_{t_{i}}$ are their corresponding ground-truth normals. $\theta_{th}$ is the angle threshold used for identifying positive and negative patches, which is empirically set to 20 degrees. 

To ensure the feature encoder network can effectively learn sharp features such as edges on the input point clouds, we choose $s_i$ and $t_i$ as close to $p_i$ as possible. To do so, we start by trying to locate $s_i$ and $t_i$ within $r$ of $p_i$, and then gradually increase the search radius if either target cannot be found. By doing so, $s_i$ is more likely to lie on the same isotropic surface as $p_i$, while $t_i$ is more likely to lie on an adjacent surface without being too far from the edge. To ensure the consistency of each triplet, the anchor, positive and negative patches within each triplet are chosen from the same point cloud.  The rotation matrix $R$ of  $\mathcal{P}_i$ is then applied to $\mathcal{S}_i$ and $\mathcal{T}_i$ as well, to ensure the network learns consistently within each triplet. 

\begin{figure}[t]
    \centering
    \includegraphics[width=\columnwidth]{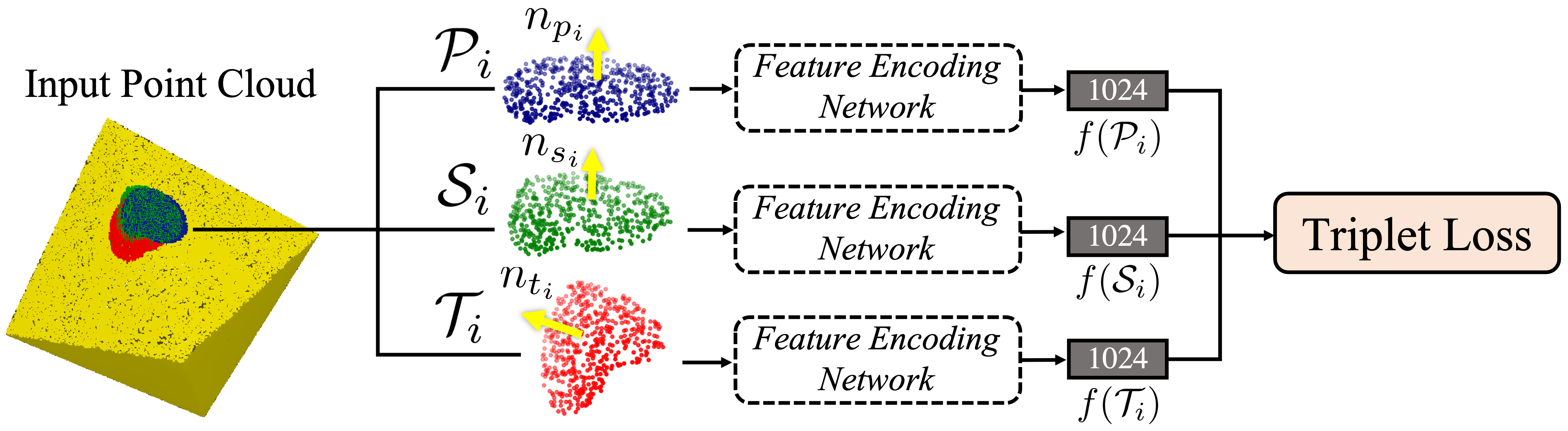}
    \caption{Illustration of triplet-based learning. The anchor, positive and negative patches are marked in dark blue, green and red respectively. 
    }
    \label{fig:upstream}
\end{figure}

\textbf{Triplet Loss}. Since input patch points are not organised in a specific order, the learning result may be affected by different permutations of the same set of points~\cite{Guerrero_pcpnet_2018}. To address this issue, we utilise an architecture which comprises of several MLPs and a max-pooling layer, inspired by PointNet~\cite{Qi_2017}, as the backbone of our encoder. Within each triplet, we perform feature extraction by feeding each patch (a 500 $\times$ 3 vector) into the network, which aggregates the features as a 1024-dimensional latent vector using the max-pooling function. We perform the same operation for $\mathcal{P}_i$, $\mathcal{S}_i$ and $\mathcal{T}_i$. The loss for the encoded triplet is given by:
\begin{equation} \label{triplet_loss_func}
    L_{E} = \mathrm{max} \{ \left \| f(\mathcal{P}_{i}) - f(\mathcal{S}_{i}) \right \|_2 - \left \| f(\mathcal{P}_{i}) - f(\mathcal{T}_{i}) \right \|_2 + m, 0 \},
\end{equation}
where $f(\mathcal{P}_{i})$, $f(\mathcal{S}_{i})$ and $f(\mathcal{T}_{i})$ are the latent representations of patch $\mathcal{P}_{i}$, $\mathcal{S}_{i}$ and $\mathcal{T}_{i}$, respectively.  Here, $f(\cdot)$ is the encoder. We empirically set the margin $m=0$. For pairwise distances of latent vectors, we use the $L_2$-norm regularisation during our training procedure. Intuitively, patches lying on isotropic and anisotropic surfaces should have small and large angles between their central point normals, respectively. Therefore, the representations of patches on an isotropic surface should be similar while patches of an anisotropic surface should be distinct. Triplet learning accomplishes this by bringing representations of patches on an isotropic surface closer while pushing apart those from an anisotropic surface, and the process is demonstrated in Fig.~\ref{fig:upstream}.
\label{label:motivation-of-triplet}

\subsection{Normal Estimation}
\label{sec:normal-estimate}

Next we aim to estimate the patch normal (i.e., the normal of the central point) from each patch's latent representation. This is accomplished by the normal estimation network which simply involves several MLPs. To train the normal estimation network, we firstly sample a patch and pre-process it in the same way as Sec.~\ref{sec:pre-process}. Then, we feed the patch into the feature encoder to obtain its 1 $\times$ 1024 latent representation which then becomes the input for the normal estimation network. The output is a 1 $\times$ 3 normal vector. 

As for the loss function for normal estimation, we define the cosine similarity between the estimated normal, $n'_{p_{i}}$, and ground-truth normals within the patch, $\{n_{p_j} | p_j \in \mathcal{P}_i\}$. We set the exponent of the cosine of the angle to 8, for the sake of preserving sharp features. Inspired by~\cite{Lu_Pointfilter_2021}, we also introduce a weighting scheme that takes into consideration the cosine similarity of the ground-truth patch's central point normal and the neighbouring ground-truth normals. Therefore, the weighted loss function $L_N$ for the normal estimation network and its weight function $\theta(n_{p_i},n_{p_j})$ are given by:
\begin{equation}
\label{eq:normal-est-loss}
L_{N} = \frac{\sum_{p_j\in \mathcal{P}_i}(1-(n'_{p_i}\cdot n_{p_j})^{8})\theta(n_{p_i},n_{p_j})}{\sum_{p_j\in \mathcal{P}_i}\theta(n_{p_i},n_{p_j})},
\end{equation}
\begin{equation}
\label{eq:weight}
\theta(n_{p_i},n_{p_j}) = \exp\left(-\frac{1-n_{p_i}\cdot n_{p_j}}{1-\cos(\sigma_s)}\right),
\end{equation}

Here, $p_j$ are points in $\mathcal{P}_i$, $p_i$ is the patch's central point, and $\theta(n_{p_i},n_{p_j})$ is the weighting function based on the ground-truth patch's central point normal and the normals of neighbouring points. $\sigma_s$ is the support angle which is set to 15 degrees by default. Since the input patch $\mathcal{P}_i$ is rotated by the rotation matrix $R$, we multiply the predicted normal by the inverse matrix $R^{-1}$ to get the final normal $\tilde{n}_{p_i}$ for the patch's central point in the original space. 

\section{Experimental Results}
\label{sec:experimental-results}

\subsection{Dataset}

\begin{figure}[t]
    \centering
    \includegraphics[width=\columnwidth]{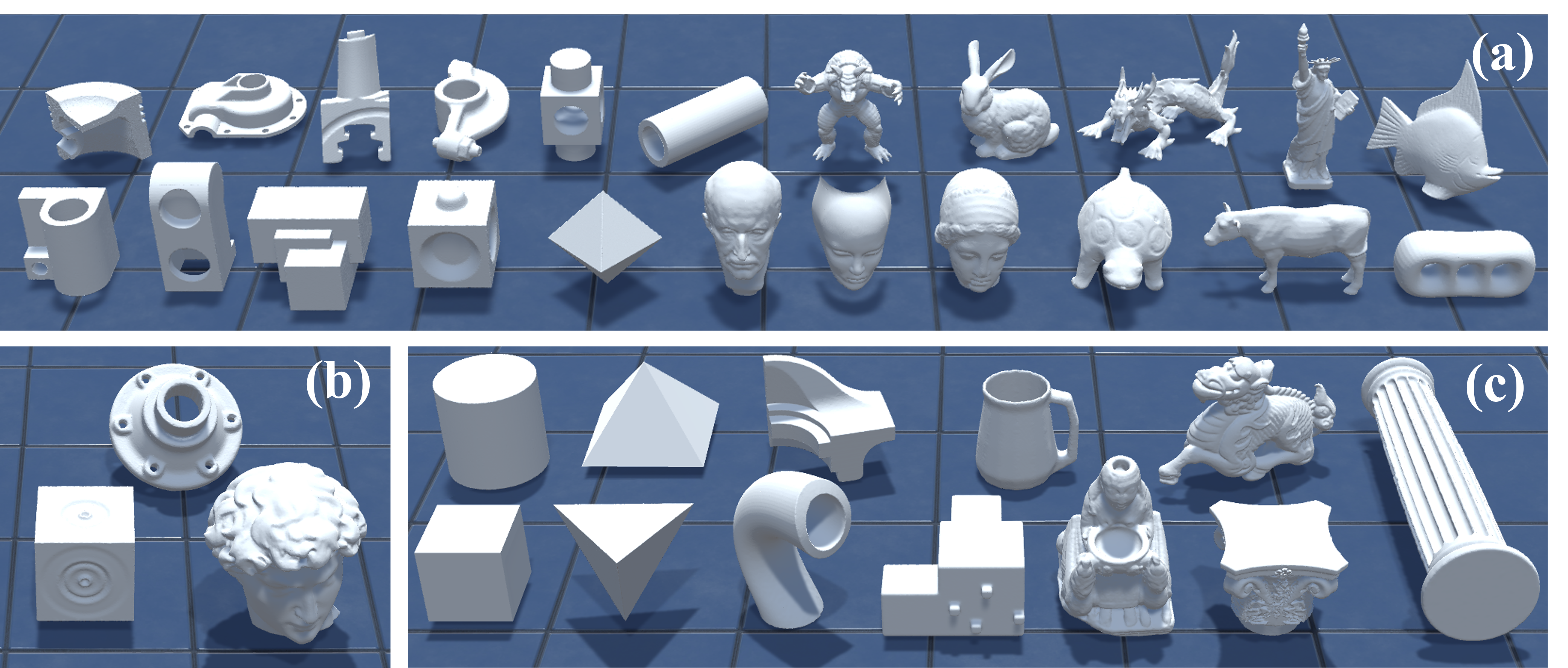}
    \caption{Demonstration of synthetic shapes where: (a) for training; (b) for validation and (c) for testing.}
    \label{train-test-shapes}
\end{figure}

Our training dataset consists of 22 clean shapes: 11 CAD shapes and 11 non-CAD shapes (Fig.~\ref{train-test-shapes}(a)). Each clean shape contains 100,000 points along with their ground-truth normals that are sampled from the shape's ground-truth surfaces. To ensure the network is trained equally on both types of shapes, we provide the same amount of training data from both CAD and non-CAD shapes. In addition, we also add noisy variants of each clean shape to the training dataset to improve the robustness of our network to noise. Each clean shape has 5 variants with different levels of Gaussian noise (0.25\%, 0.5\%, 1\%, 1.5\% and 2.5\% of each clean shape's bounding box diagonal length, respectively). Therefore, we have 132 (22 $\times$ 6) training shapes in total.

In both the feature encoding and normal estimation training stages, we sample 8,000 patches from each shape to ensure the network generalises well on sufficient data. We validate our network model on the validation set (Fig.~\ref{train-test-shapes}(b)), and test on 12 synthetic shapes (Fig.~\ref{train-test-shapes}(c)) with various noise levels, as well as additional raw scanned point clouds. 





\begin{figure}[h!]
    \centering
    \includegraphics[width=\columnwidth]{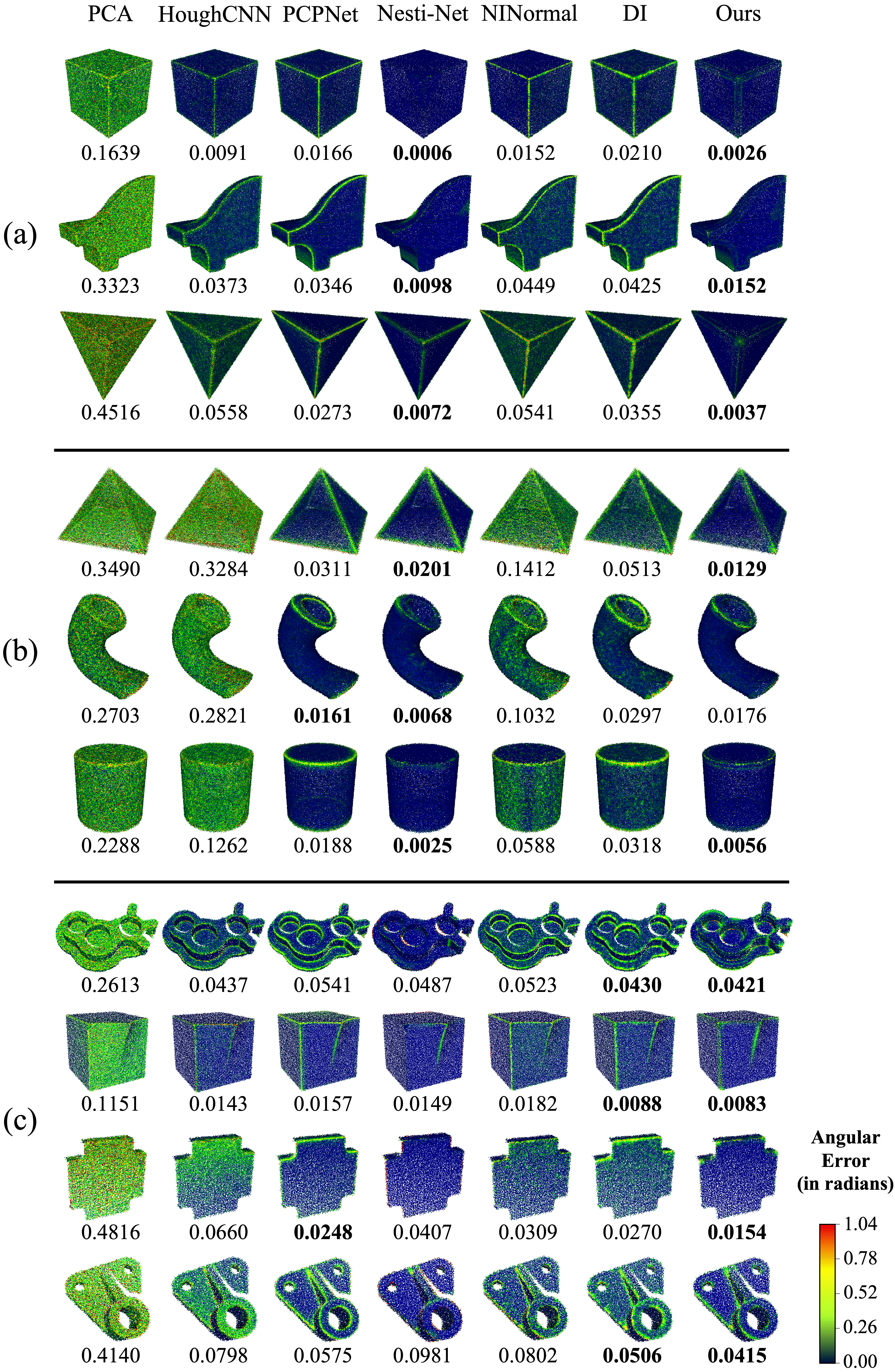}
    \caption{MSAE comparisons on \textit{noisy} CAD shapes, where (a) and (b) demonstrate MSAE on synthetic shapes, and (c) demonstrates MSAE on scanned shapes with known ground-truth normals. The two best results are in bold in each row.}
    \label{fig:noisy-cad-msae}
\end{figure}

\subsection{Implementation Details}
We implement our networks using PyTorch 1.8.0, train and test them on an NVIDIA GeForce RTX 3080 GPU with CUDA 11.3. Both feature encoding and normal estimation networks are trained using an SGD optimiser, with a momentum of 0.9 and an initial learning rate of 0.01. We train the feature encoding network for 5 epochs since it converges quickly, and train the normal estimation network for 50 epochs. During training, the learning rate is multiplied by a 0.1 factor if no improvement happens within 3 consecutive epochs.

\begin{table*}[t]
\begin{center}
\caption{Average MSAE of each method on the testing shapes.} \label{tab:avg-msae}
\scalebox{0.95}{
\begin{tabular}{|c|c|c|c|c|c|c|c|}
\hline
Method  & PCA             & HoughCNN & PCPNet & Nesti-Net & NINormal & DI & Ours            \\ \hline
CAD Shapes     & 0.3068          & 0.1043   & 0.0297 & 0.0249    & 0.0599 & 0.0341  & \textbf{0.0165} \\
Non-CAD Shapes & 0.4582 & 0.2747   & \textbf{0.2439} & 0.2539    & 0.2510 & 1.0535   &    0.2636          \\ \hline
Overall & 0.3636 & 0.1682   & 0.1100 & 0.1108    & 0.1316 & 0.4164   &    \textbf{0.1092}          \\ \hline
\end{tabular}
}
\end{center}
\end{table*}

\begin{table*}[t]
\begin{center}
\caption{Network size comparison among different approaches and inference time comparison (seconds per 100K points).} \label{tab:modelsize-and-time}
\scalebox{0.95}{
\begin{tabular}{|c|c|c|c|c|c|c|c|}
\hline
Method Name        & HoughCNN & PCPNet & Nesti-Net & NINormal & DI & DFP & Ours           \\ \hline
Network Size (in MB) & 111.6 & 85.4   & 2020.0      & 39.5  & \textbf{0.037}   & 268.4 &  10.42 \\
Inference Time (in Seconds) & 74.67 & 227.33   & 1234.5  & \textbf{2.7}  &  3.7 & 60840.0 & 55.6 \\ \hline
\end{tabular}
}
\end{center}
\end{table*}


\begin{table}[t]
\begin{center}
\caption{The MSAE of the shapes in our validation set regarding the choice of the exponent in Eq.~\eqref{eq:normal-est-loss}.} \label{tab:ablation-power}
\scalebox{0.95}{
\begin{tabular}{|c|c|c|c|c|}
\hline
Exponent & 2               & 4      & 8               & 12     \\ \hline
Box-Groove       & 0.0149          & 0.0101 & \textbf{0.0097} & 0.0145 \\
Carter       & \textbf{0.1285} & 0.1320 & 0.1302          & 0.1472 \\
David       & 0.1040          & \textbf{0.1031} & \textbf{0.1031} & 0.1279 \\ \hline
Average      & 0.0825          & 0.0817 & \textbf{0.0810} & 0.0965 \\ \hline
\end{tabular}
}
\end{center}
\end{table}

\begin{figure}[h]
    \centering
    \includegraphics[width=0.8\columnwidth]{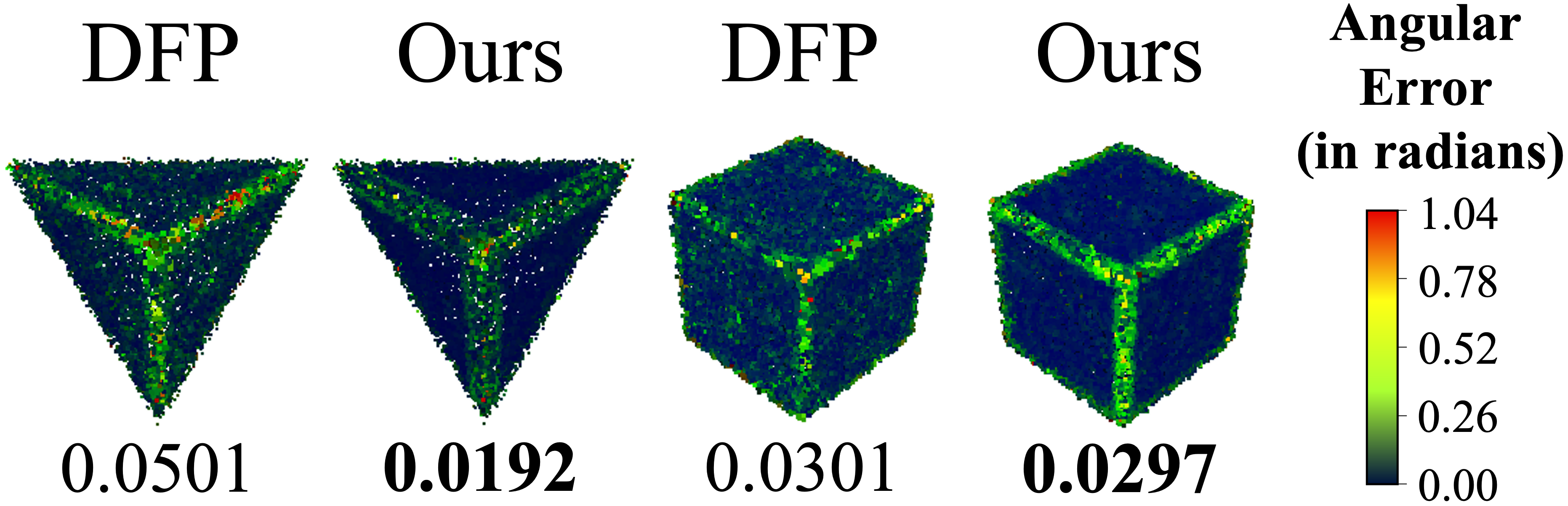}
    \caption{MSAE comparisons with DFP on Tetrahedron and Cube with 1\% noise.
    }
    \label{fig:dfp-compare}
\end{figure}

\subsection{Comparisons}
We compare our normal estimation results on the test dataset with PCA~\cite{Hoppe_1992}, HoughCNN~\cite{Boulch_Hough_2016}, PCPNet~\cite{Guerrero_pcpnet_2018}, Nesti-Net~\cite{Benshabat_Nesti_2019}, NINormal~\cite{Wang_NINormal_2020}, DI~\cite{Lenssen_DI_2020}, and additionally, DFP~\cite{Lu_DFP_2020}. During the comparison study, we retrained all methods, except for HoughCNN and DFP, on our training dataset and used recommended values of parameters for each respective method during testing, to ensure fair comparisons. HoughCNN creates synthetic point set data and computes corresponding accumulators while DFP has a point classification step requiring labelled ground-truth points, indicating they cannot be trained on our dataset. 
Also, methods such as~\cite{Boulch_Hough_2016, Benshabat_Nesti_2019, Wang_NINormal_2020} sometimes produce normals with wrong orientations, and the results are not normalised. Therefore, we flip and normalise the affected normals to guarantee fair comparisons. We use mean squared angular error (MSAE)~\cite{Lu_DFP_2020} for evaluating the accuracy of the predicted normals against the ground-truth ones. Table~\ref{tab:avg-msae} shows MSAE values for the different methods over the testing shapes, where our method achieves the minimum overall MSAE.

\textbf{Synthetic Point Clouds.} Fig.~\ref{fig:noisy-cad-msae}(a) and Fig.~\ref{fig:noisy-cad-msae}(b) demonstrate results on 6 CAD shapes with 0.5\% and 1\% random vertex displacement noise, added using MeshLab~\cite{meshlab}, respectively. Our method outperforms all other methods on Tetrahedron and Pyramid (the third and fourth rows), and ranks the second on Cube, Fandisk and Cylinder (the first, second and sixth rows). While our method generates outputs with less obvious gains on non-CAD shapes since it tends to over-sharpen features, it still produces the minimum MSAE among all testing shapes overall, as shown in Table~\ref{tab:avg-msae}. 

We also demonstrate comparison results with DFP~\cite{Lu_DFP_2020} which has two separately trained models for feature points and non-feature points. For fair comparisons, we use normal estimation results from the first iteration of DFP since it optimises point positions in further iterations. As can be seen from Fig.~\ref{fig:dfp-compare}, our method outperforms DFP on 2 noisy CAD shapes with 1\% random vertex displacement noise in terms of MSAE and inference time: DFP takes 1.69 hours to process every 10,000 points, while ours only requires 5.56 seconds. Since the two shapes in Fig.~\ref{fig:dfp-compare} have a much smaller number of points in them (10,171 points for Tetrahedron and 16,747 points for Cube), it also demonstrates the robustness of our method on less dense point clouds.

\label{sec:scanned-models}
\textbf{Scanned Point Clouds.} We also test the effectiveness and scalability of our method on scanned point clouds, which are generally incomplete and consist of uneven point distributions compared to synthetic ones. Fig.~\ref{fig:noisy-cad-msae}(c) demonstrates comparison results on 4 raw scanned point clouds of CAD shapes, where the points are corrupted by noise and their ground-truth normals are known. As shown in Fig.~\ref{fig:noisy-cad-msae}(c), most methods' results are suboptimal; while Nesti-Net performs relatively well on flat surfaces, there are many predicted normals at the edges that point to wrong directions, increasing the overall angular error. Among all methods, our approach achieves the minimum MSAE on the 4 raw scanned shapes.

\textbf{Network Size and Inference Time.} It is worth noting that our method achieves the state-of-the-art performance with a relatively small network size and estimates normals at a relatively faster speed. Table~\ref{tab:modelsize-and-time} lists the comparison of all network sizes and inference time for all methods. Although Nesti-Net slightly outperforms our results on 4 CAD shapes listed in Fig.~\ref{fig:noisy-cad-msae}, it utilises the largest network. The most lightweight method, DI, and the fastest method, NINormal, both show high sensitivity to noise.

In addition, our method is more robust to irregular sampling and fewer points. \textit{Please refer to our supplementary document for additional results.}

\subsection{Ablation Study}

\textbf{Normal Estimation.} During the normal estimation phase, we introduce a cosine similarity exponent of 8 in Eq.~\eqref{eq:normal-est-loss} as it effectively preserves sharp features on noisy shapes compared to others. We evaluate this over our validation set 
using different exponents. As shown in Table~\ref{tab:ablation-power}, the average MSAE value is the minimum when the exponent is set to 8.


\begin{figure}[t]
\centering
\includegraphics[width=0.9\columnwidth]{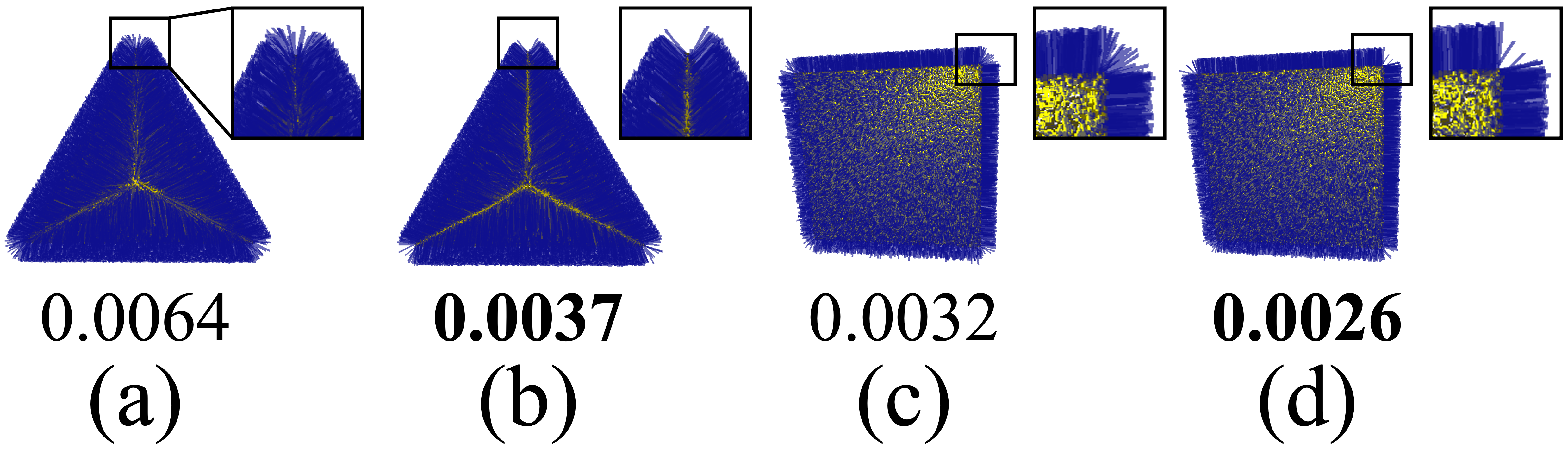}
\caption{Visualised normals and MSAE (a, c) without, and (b, d) with triplet learning-based feature encoding network.} 
\label{fig:ablation-edge}
\end{figure}

\textbf{Feature Encoding Network.} Alternatively, we can directly regress the normal vector without our feature encoding network. As illustrated in Fig.~\ref{fig:ablation-edge}, utilising triplet loss to train our feature encoding network is effective in maintaining sharp edges and thus achieves a smaller MSAE value.

\section{Conclusion}
\label{sec:conclusion}
In this paper, we presented a novel deep learning normal estimation method for 3D point clouds. In its feature encoding phase, the feature encoder creates latent representations of local patches through optimising relative distances within triplets. In the normal estimation phase, these representations are consumed by the normal estimation network which regresses the normals for the central points in the patches using MLPs. Comparison results with other representative methods have demonstrated that our method has achieved the state-of-the-art performance in estimating normals in the presence of noise, especially for sharp features.

\bibliographystyle{IEEEbib}
\bibliography{icme2022template}

\end{document}